\documentclass[
]{ceurart}

\usepackage{listings}
\usepackage{algorithm}
\usepackage{algpseudocode}
\usepackage{subcaption}
\usepackage{wrapfig}
\usepackage{float}

\begin{document}

\copyrightyear{2022}
\copyrightclause{Copyright for this paper by its authors.
  Use permitted under Creative Commons License Attribution 4.0
  International (CC BY 4.0).}


\conference{Accepted at NeSy 2022, 16th International Workshop on Neural-Symbolic Learning and Reasoning, Cumberland Lodge, Windsor, UK}

\title{ Knowledge-based Analogical Reasoning in Neuro-symbolic Latent Spaces  }



\author[1]{Vishwa Shah}[%
email=f20180109@goa.bits-pilani.ac.in,]
\author[1]{Aditya Sharma}
\author[2]{Gautam Shroff}
\author[2]{Lovekesh Vig}
\author[1]{Tirtharaj Dash}
\author[1]{Ashwin Srinivasan}

\address[1]{APPCAIR, BITS Pilani, K.K. Birla Goa Campus} \address[2]{TCS Research, New Delhi}




\begin{abstract}
 Analogical Reasoning problems pose unique challenges for both connectionist and symbolic AI systems as these entail a carefully crafted solution combining background knowledge, deductive reasoning and visual pattern recognition. While symbolic systems are designed to ingest explicit domain knowledge and perform deductive reasoning, they are sensitive to noise and require inputs be mapped to a predetermined set of symbolic features. Connectionist systems on the other hand are able to directly ingest rich input spaces such as images, text or speech and can perform robust  pattern recognition even with noisy inputs. However connectionist models struggle to incorporate explicit domain knowledge and perform deductive reasoning. In this paper, we propose a framework that combines the pattern recognition capabilities of neural networks with symbolic reasoning and background knowledge for solving a class of  Analogical Reasoning problems where the set of example attributes and possible relations across them are known apriori. We take inspiration from the `neural algorithmic reasoning' approach [DeepMind 2020] and exploit problem-specific background knowledge by  (i) learning a distributed representation based on a symbolic model of the current problem (ii) training neural-network transformations reflective of the relations involved in the  problem and finally (iii) training a neural network encoder from images to the distributed representation in (i). These three elements enable us to perform search-based reasoning using neural networks as elementary functions manipulating distributed representations. We test our approach on visual analogy problems in RAVENs Progressive Matrices, and achieve accuracy competitive with human performance and, in certain cases, superior to initial end-to-end neural-network based approaches. While recent neural models trained at scale currently yield the overall SOTA, we submit that our novel neuro-symbolic reasoning approach is a promising direction for this problem, and is arguably more general, especially for problems where sufficient domain knowledge is available.

\end{abstract}



\maketitle

\section{Introduction}
Many symbolic reasoning algorithms can be viewed as searching for a solution in a space defined by prior domain knowledge. Given sufficient domain knowledge represented in symbolic form, `difficult' reasoning problems, such as analogical reasoning, can be `solved' via exhaustive search, even though they are challenging for the average human. However, such algorithms operate on a symbolic space, whereas humans are easily able to consume rich data such as images or speech. Neural networks on the other end are proficient in encoding high-dimensional continuous data and are robust to noisy inputs, but struggle with deductive reasoning and absorbing explicit domain knowledge. `Neural Algorithmic Reasoning'\cite{velivckovic2021neural}, presents an approach to jointly model neural and symbolic learning, wherein rich inputs are encoded
to a latent representation that has been learned using from symbolic inputs.
This design allows neural learners and algorithms to complement each other's weaknesses. Through this work, we aim to investigate a variation of the neural algorithmic reasoning approach applied to analogical reasoning, using  RAVENs Progressive Matrices \cite{Carpenter} problems as a test case.

\begin{wrapfigure}{r}{0.45\textwidth}
    \centering
    \includegraphics[scale=0.3]{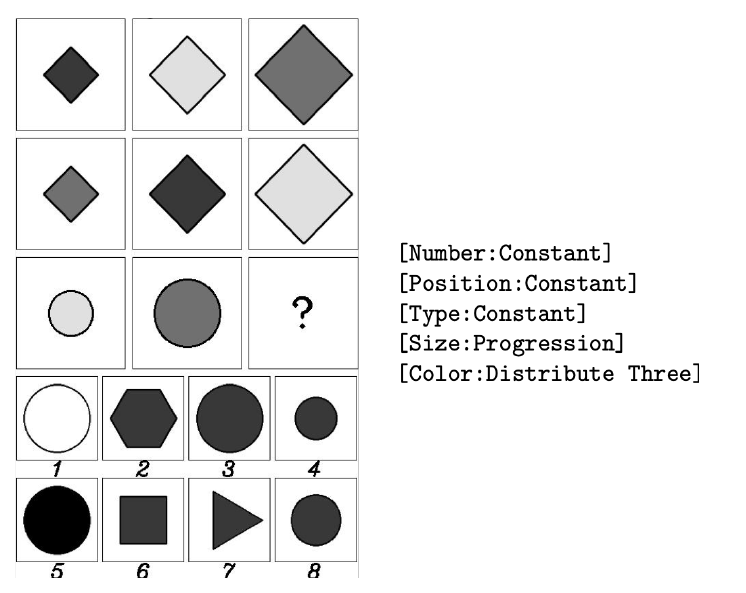}
    \captionof{figure}{RPM : Problem Matrix (Top), Answer Options (Bottom)}
    \label{example_rpm}
\end{wrapfigure}

Our approach essentially exploits the domain knowledge to train a suite of neural networks, one for each known domain predicate. For
RAVENs problems, these predicates represent rules that might apply to ordered sets (rows) of images in a particular problem. Further,
these neural predicates are trained to operate on a special high-dimensional representation space (`symbolic latent space') that is derived, via self-supervised learning, from the symbolic input space. Note that a purely symbolic algorithm can consume symbolic inputs to solve the problem exactly, however a distributed representation can allow for real world analogical reasoning for rich input spaces such as images or speech (see Fig \ref{example_rpm} for a RAVENs problem; one can also imagine tasks with speech inputs where the analogous example signals are high pitch transformed versions of the original). Our approach differs from \cite{velivckovic2021neural} where the symbolic latent space is derived via a supervised approach; by using a self-supervised learning approach we are able to use the same representation space to train multiple neural predicates, unlike in \cite{velivckovic2021neural} where only a single function is learned. Next, we train a neural network encoder to map real-world images (here sub-images of the RAVENs matrices)
to the `symbolic latent space'. Finally, using the above elements together we are able to perform symbolic search-based reasoning, albeit using neural-networks as primitive predicates, to solve analogical reasoning problems. 

\noindent{\textbf{Contributions}} (1) We adapt and extend Neural Algorithmic Reasoning to propose a neuro-symbolic approach for a class of visual analogical reasoning problems 
(2) We present experimental results on the RAVENs Progressive Matrices dataset and compare our neuro-symbolic approach to purely connectionist approaches, and
analyse the results. In certain cases, our approach is superior to initial neural approaches, as well as to human performance (though more recent neural approaches
trained at scale remain SOTA) (3) We submit that our approach can be viewed as a novel and more general neuro-symbolic procedure that uses domain knowledge to 
train neural network predicates operating on a special, `symbolically-derived latent space', which are then used as elementary predicates in a symbolic search process.

\section{Problem Definition}
\label{sec:problem}
In general, `RAVEN-like' analogical reasoning tasks can be viewed as comprising of $n$ ordered sets $s_1, s_2,..., s_n$ containing $m$ input samples each, an additional test set containing $m-1$ samples and a target $m^{th}$ sample. Each sample $I_{jk}$ where $j \in [1...n], i \in [1...m] $ is comprised of a set of entities $E_{jk}$ and each entity $e \in E_{jk}$    has attributes from a set $A$ of $k$ predefined attributes ${a_1, a_2,..., a_k} \in A$. Assume a predefined set of all possible rules $R={r_1, r_2...,r_u}$ that can hold over sample attributes in an example set(s). For a given task the objective is to infer  which rules hold across the $m$ samples in each of the $n$ example sets in order to subsequently predict the analogous missing sample for the test set, either by generating the target sample as in the ARC challenge \cite{chollet2019measure}, or by classifying from a set of possible choices as in RPMs. Note that the problem definition assumes prior domain knowledge about possible sample entities, their possible attribute values, and possible rules over sample attributes within the example sets. It is worth mentioning that while here we investigate visual analogies, the problem definition can accommodate input samples of any datatype such as audio or text as long as the problem structure is unchanged.

\section{Proposed Approach}
\label{sec:approach}

\begin{center}
\includegraphics[scale=0.35]{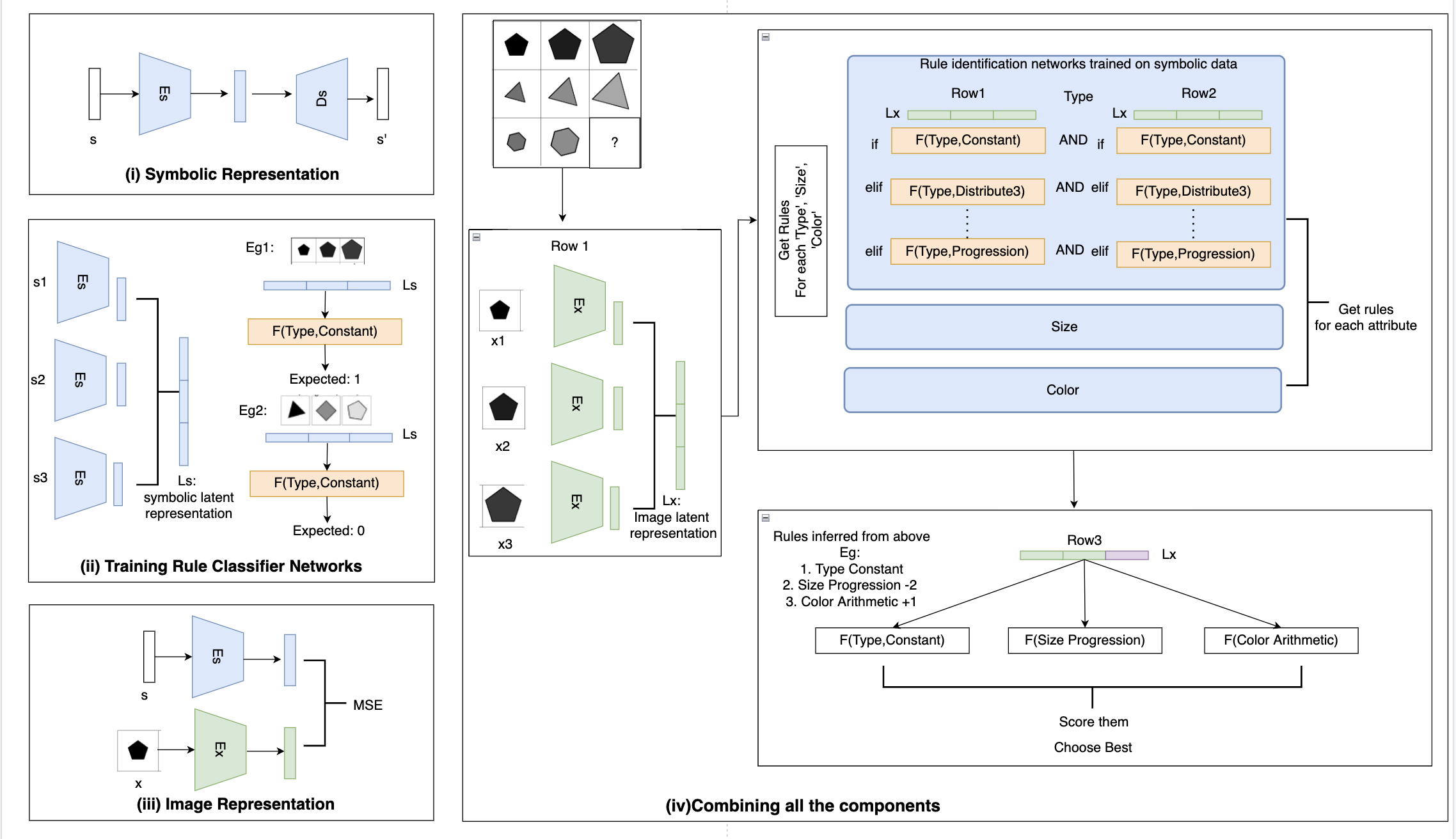}
\captionof{figure}{Overview of our approach for RAVEN's RPM}
\label{pipeline}
\end{center}

We adapt a variation of the neural algorithmic reasoning approach to the problem defined in Section \ref{sec:problem}, where we (i) learn a latent representation based on the symbolic representation of the tasks, via self-supervised learning; (ii) train neural networks to infer the rules involved in the problem; (iii) train a neural encoder from images to align with the symbolic latent representations in (i), and (iv) use the above
elements to solve a given task via a neuro-symbolic search procedure, i.e.,
where the elementary predicates are neural networks. We assume the presence of a training dataset $D_{train}$ with correct answer labels for the analogical reasoning tasks. The components (i), (ii) and (iii) are trained independently as for each of them we know or can determine the inputs and the targets depending on their function. We also evaluate an alternative of (v) encoding an image
to the symbolic latent space via the neural encoder in (iii) above, and decode it to symbolic form using the decoder trained in (i) on the symbolic space,
on which purely symbolic search is used to solve a problem instance.

\subsection{Learning a Distributed Representation from the Symbolic Space}
\label{sec:symbolic}
We begin with a symbolic multihot task representation $s$, which is a series of concatenated one-hots, one for each image entity attribute. Each attribute can take a value from a finite set and hence is represented using a one-hot vector. To obtain the latent representations of the tasks, we train an auto-encoder on the symbolic task definitions $\mathbf{S}$ as $\mathbb{L}(\mathbf{S}) = (E^{\mathbf{S}}_{\theta},D^{\mathbf{S}}_{\phi})$ where the encoder $E^{\mathbf{S}}_{\theta}$ maps from the symbolic space to the latent space and the decoder $D^{\mathbf{S}}_{\phi}$ maps the representation from the latent space back to the symbolic space as shown in component (i) of Fig. \ref{pipeline}. As we want to reconstruct the multihot representation, a sum of negative-log likelihood is computed for each one-hot encoding present in the multihot representation. We provide an example in \ref{sec:autoencoder_loss} where the parameters $\theta$ and $\phi$ are obtained via gradient descent on a combination of negative log-likelihood loss functions as shown in equation \ref{loss} and \ref{params} in the appendix.

\subsection{Training Rule Identification Neural Networks}
\label{predicates}
Next for every attribute, and for each applicable rule for that attribute, we train a Rule Identification neural network classifier to predict if the rule (pattern) holds for the example set. As mentioned in \ref{sec:problem}, we know the rules that can hold over attributes, giving us a definite set of networks to be trained. We refer to any rule identification network $F$ using the $(attribute, rule-type)$ pair for which it is trained.
The latent representations obtained after encoding the symbolic representations of each of the samples in the example set are concatenated and sent as input to the rule identification networks. While training a neural net for a $(attribute,rule-type)$ pair, we categorize each example set with the specific label for that particular rule and attribute, labeling it with 0 if the rule-type is not followed, 1 if the rule-type is followed or a rule-value indicating the level of the rule-type when being followed.  As each of these rule-types is deterministic, we can obtain the rule-type and value for each row using their symbolic representations. The overview can be seen in component (ii) of Fig. \ref{pipeline} where we see the input for these elementary neural networks and the expected output determined for the $(attribute, rule-type)$ pair. We see in Fig. \ref{pipeline} that type (shape) changes in row $Eg1$, hence the expected target for $F(type,constant)$ should be 1 as type stays constant and in case of $Eg2$ as the type changes, we expect $F(type,constant)$ to predict 0, indicating the rule is not obeyed. With these labels, each network is optimized using cross-entropy loss. For parameterized rules we train additional networks to predict the parameters.\footnote{Examples provided in appendix section \ref{C.2}}

\subsection{Sample representation}
\label{sec:image}
As our objective to apply our approach on samples in an unstructured (image, text, speech) format, we want to develop a representation for the samples that resembles the latent representation of symbolic inputs. For this we train any neural network encoder $E^{\mathbf{X}}_{\psi}$ over the rich input space $\mathbf{X}$ which encodes each sample to a latent space. We want to minimize the disparity in the latent representations from the sample $x$ and its corresponding symbolic representation $s$. For this we use $E^\mathbf{S}_\theta$ trained in \ref{sec:symbolic}. We minimize the mean squared error over all pairs of symbolic and input representations (equation shown in \ref{sec:img_encoder_los}). This enables us to use the previously learned neural networks for rule inference on our image data.

\subsection{Combining the Elements}
As shown in component (iv) of Fig. \ref{pipeline}, we first use $E^{\mathbf{X}}_{\psi}(x)$ as inputs to find the underlying rules using the neural networks trained in section \ref{predicates}. Once we obtain the set of (rule-type, value) pairs for each attribute, we apply these neural networks for each of the answer choices by adding them to the test set. For each attribute, we obtain the output probability score for that rule-type and value. The final score is the sum of these probability scores for all the inferred rules \footnote{We explain the complete pseudo-algorithm along with the scoring function in the appendix.}. The choice with the highest score is returned as the answer.


\section{Empirical Evaluation}

\textbf{Raven's Progressive Matrices (RPM):} is a widely accepted visual reasoning puzzle used to test human intelligence \cite{Carpenter}. The idea is to infer multiple patterns to be able to find the answer from the options, an example of the same from the RAVEN\cite{zhang2019raven} dataset is seen in Fig \ref{example_rpm}. In this dataset, for every problem, each independent attribute follows an algorithmic rule. The task here is to deduce the underlying rules applied over each row for the first two rows; followed by selecting the option that validates all the inferred rules when aligned with the images in the last row. As seen in the first two rows in the Fig \ref{example_rpm} we observe the attributes: Number, Position and Type stay constant across the rows. We observe an increasing progression in Size and a fixed set of 3 Colors are permuted within the row indicating distribute three. Hence option 3 is the only solution that satisfies all the rules\footnote{We provide the dataset overview, set of rule and attribute definitions for the RAVENs problems used in the appendix.}. For our experiments, We use the ``Relational and Analogical Visual rEasoNing'' dataset (RAVEN)\cite{zhang2019raven}, which was introduced to drive reasoning ability in current models. RAVEN consists of 10,000 problems for each of the 7 different configurations of the RPM problem shown in Fig \ref{7-configs}. Each problem has 16 images (8 : problem matrix and 8 : target options).

\begin{center}
\includegraphics[scale=0.20]{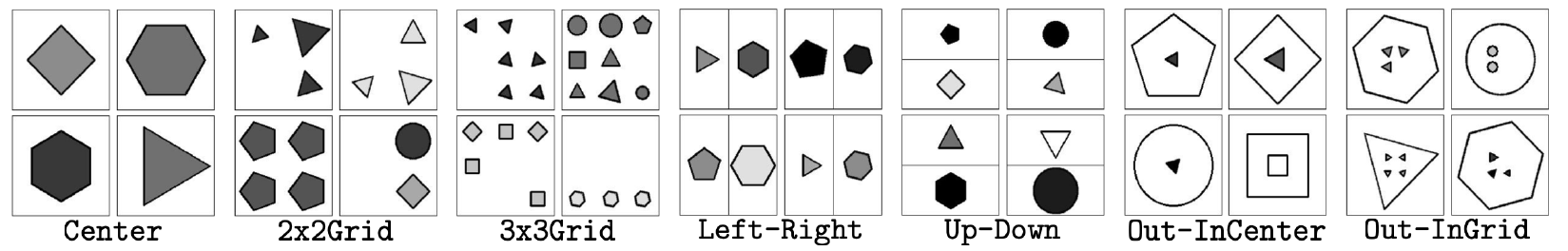}
\captionof{figure}{Examples of 7 different configurations of the RAVEN dataset}
\label{7-configs}
\end{center}

\subsection{Experimental Details}
As each image in a RAVENs problem can be represented symbolically
in terms of the entities (shapes) involved and their attributes: Type, Size, and Color; and multiple entities in the same image have Number and Position attributes. Such attributes are also rule-governing in that rules 
based on these can be applied to each row 
and the combination of rules from all rows is used to solve a given
problem.
Example: for each entity, the multihot representation $s$ is of size  $|T| +|S|+|C|$ where $T$, $S$ and $C$ are the set of shapes, possible sizes and possible colors respectively. The multi-hot vector is made up of 3 concatenated one-hot vectors, one each for type, size, and color. In case of multiple components, e.g: Left-Right, we concatenate the multihots of both the entities. For our auto-encoder architecture, we train simple MLPs with a single hidden layer for both $E_s$ and $D_s$ (using loss function \ref{loss},\ref{params}) in appendix. The dimensions of the layers and latent representations are chosen based on the RPM configuration.

Following \cite{Carpenter}'s description of RPM, there are four types of rules: Constant, Distribute Three, Arithmetic, and Progression. In a given problem,
there is a one rule applied to each rule-governing attribute across all rows,
and the answer image is chosen so that this holds.
We aim to find a set of rules being obeyed by both the rows. 

So for every attribute and its rule-type, we train an elementary neural network classifier $F_{(attribute,rule-type)}$, to verify if the rule is satisfied in a given row, or pair of rows. The rules of Progression and Arithmetic are further associated with a value (e.g., Progression could have increments or decrements of 1 or 2). For rule-type Constant and `Distribute-three' we train a binary classifier, and for rule-type Arithmetic and Progression, we train a multi-class classifier to also predict the value associated with the rule. An example is described in the appendix along
with further details on the neural networks used.

 We train a CNN-based image encoder $E^{\mathbf{X}}_{\psi}$ over the image space $\mathbf{X}$ which encodes each image of the problem to a latent space and minimize the disparity with the corresponding symbolic latent space as described in Section \ref{sec:image}. Finally, as shown in component (iv) of Fig. \ref{pipeline}, we find the underlying rules using the neural networks trained in section \ref{predicates}. Once we obtain the set of (rule-type, value) pairs for each attribute, we apply these neural networks for each of the 8 options by placing them in the last row. We obtain the output probability score for that attribute, rule-type and value and sum the probability scores for all the inferred rules \footnote{We explain the complete pseudo-algorithm along with the scoring function in the appendix.} and the image with the highest score is returned as the answer.

\subsection{Results}

\begin{table*}[]
\caption{Configuration Wise Accuracy}
\label{Configuration_Wise_Accuracy}
\scriptsize
\begin{tabular}{@{}cccccccc@{}}
\toprule
                  & Center & Left-Right & Up-Down & Out-In Center & 2x2 Grid & 3x3 Grid & Out-In Grid \\
    Input/Reasoning
                \\ \midrule

\textbf{A: Image/Neural(ours)}    &  89.40\%  &       85.00\%    &  89.10 \%   &   89.80\%            &  53.10\%   &  33.90\%         &   31.90\%          \\
\textbf{B: Image/Symbolic(ours)}    &  97.30\%  &    98.35\%    &  98.95\%   &   96.95\%            &  88.40\%   &  19.15\%         &   34.15\%          \\
\textbf{c: Symbolic/Neural(ours)} &  94.60\%  &    90.65\%    &  91.90\%   &   93.85\%            &  62.20\%   & 54.10\%         &   59.40\%          \\ \midrule

RAVEN(ResNet+DRT)\cite{zhang2019raven} & 58.08\% & 65.82\% & 67.11\% & 69.09\% & 46.53\% & 50.40\% & 60.11\% \\
CoPINet\cite{zhang2019learning} & 95.05\%  & 99.10\% & 99.65\% & 98.50\% &  77.45\% & 78.85\% & 91.35\%
\\
SCL\cite{SCL}  & 98.10\%   & 96.80\% & 96.50 \% & 96.00\% & 91.00\%  & 82.50\% & 80.10\% \\
DCNet\cite{DCNet}  &  97.80\%  &  99.75\% &  99.75\% & 98.95\% &   81.70\% & 86.65\% & 91.45\% \\
\midrule
Human \cite{zhang2019raven} &  95.45\% &  86.36\% & 81.81\% & 86.36\% & 81.82\% & 79.55\% & 81.81\% \\

\bottomrule
\end{tabular}
\end{table*}



We use the test set provided by RAVEN to evaluate rule classification networks and the final accuracy.  Table \ref{F1-score_classification_networks} lists the F1 of each $F_{(attribute,rule-type)}$ classification network across all configurations. We observe that $85\%$ of the neural networks have an F1-score $\ge$ 0.90. This is corroborated by the idea that these networks are trained on latent representations of symbolic data to perform elementary functions and do well on specialized reasoning components.

Table \ref{Configuration_Wise_Accuracy} shows end-to-end accuracy for different RAVENs problem configurations. Our proposed neural reasoning approach
(\textbf{A})
is where we have \textbf{image} input encoded by $E_x(x)$
and \textbf{neural} reasoning in
the latent space, i.e. steps (i)-(iv) in Section \ref{sec:approach}. We
also show results for an alternative (\textbf{B}), (v) mentioned in
Section \ref{sec:approach}, i.e.,
\textbf{image} inputs decoded to symbolic space via $D_s(E_x(x))$
followed by purely \textbf{symbolic} reasoning (algorithmic solving).
Results using \textbf{neural} reasoning in the latent space
but using the correct \textbf{symbolic} inputs
mapped via $E_s(s)$ are
also shown as (\textbf{c}) to highlight the loss in accuracy incurred while encoding images
using $E_x(x)$.

We use ResNet+DRT from RAVEN as our baseline, human performance (provided in \cite{zhang2019raven} ) as a reference and other SOTA methods for comparison. We note that the RAVEN
baseline is bested by \textbf{A}: neural reasoning on image inputs for 4 out of the 7 configurations, and by \textbf{B}: symbolic reasoning on image inputs
for one of the more difficult cases (2x2). At the same time we observe that approach
\textbf{B} is better than \textbf{A} except for the difficult case of 3x3 grid,
where the encoder-decoder combination $D_s(E_x(x))$ produces too many
errors, adversely affecting purely symbolic reasoning.

Neural reasoning from symbolic inputs, i.e. (\textbf{c}), accuracy consistently exceeds approach \textbf{A}, which can be attributed to the closer relation of the latent space to the algorithmic symbolic space.
We also observe lower performance for the configurations 2x2 Grid, 3x3 Grid, and Out-In grid. Upon analysis, we find that the performance of $E_x$ for these configurations is relatively lower as each of these components have multiple entities and the task to transform the image into the latent space and identifying rules becomes difficult.

While more recent purely neural-network based approaches remain SOTA, we note that for the simpler configurations our neuro-symbolic approaches are competitive.
We speculate that because of the complex nature and difficulty of these configurations, using more powerful neural architectures (such as transformers) for self-supervised learning
of the symbolic latent space as well as for learning predicates can be useful. More generally our results provide evidence that a neuro-symbolic search using
neural-network based elementary predicates, trained on a symbolic latent space, may
be a promising approach for learning complex reasoning tasks, especially where
domain knowledge is available.

\section{Discussion}
While the results presented in this paper pertain to visual analogical reasoning problems, it should be noted that the procedure presented in Section 3 is agnostic to the input modality. Figure \ref{fig:generalization} illustrates  analogical reasoning problems in speech and text respectively; the first task  involves analogical reasoning in speech, where the input corresponds to a speech sample in a male voice and the output samples correspond to the same utterance in a female voice: The task is to infer that this is the transformation involved and analogously generate the output speech signal for the target query. Possible attributes for rules on a speech signal can include discrete values of pitch, tone, volume or others. In the second, text-based example, inputs correspond to positive reflections of an input passage, and the outputs correspond to negative reflections of the same passage. Attributes for text rule identification can similarly include textual aspects like language, sentiment and style. Note that both these examples require generation of the missing target output which is a harder task than classification from a set of possible choices. However, given the recent progress in conditional generation for images \cite{Dalle} and text\cite{Pnp}, this seems entirely feasible.

\begin{center}
\includegraphics[scale=0.15]{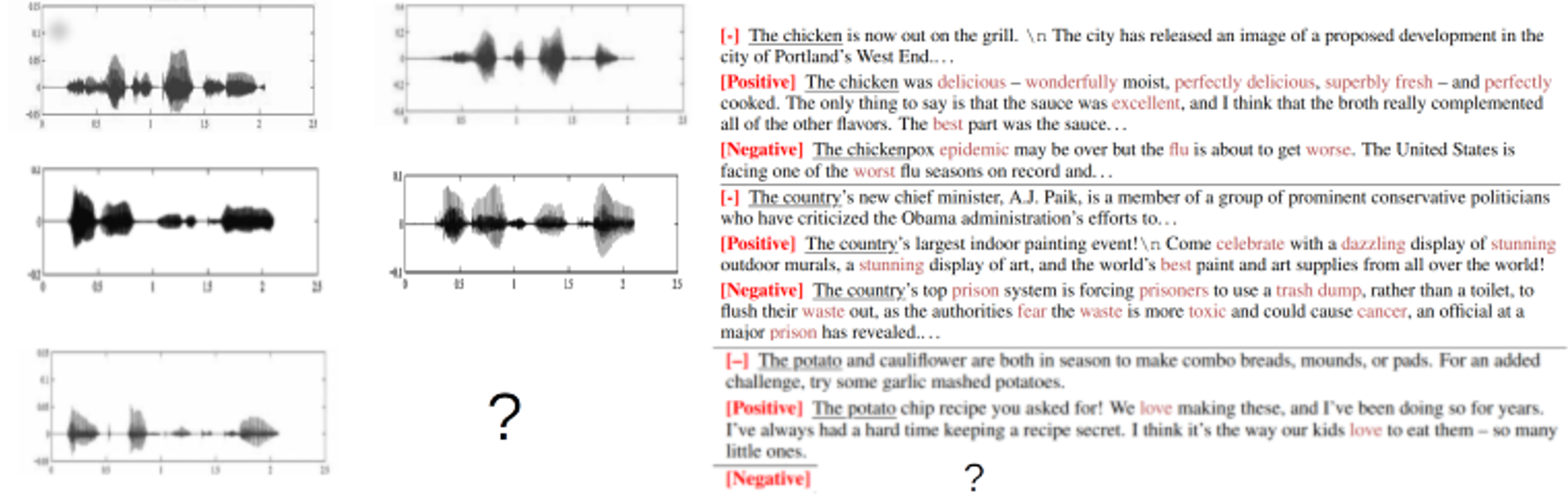}
\captionof{figure}{Analogical reasoning problems across different input modalities.}
\label{fig:generalization}
\end{center}

\section{Related Work}

\noindent The `neural algorithmic reasoning'\cite{velivckovic2021neural} approach presents a procedure for building neural networks that can mimic algorithms. It includes training processor networks that can operate over high-dimensional latent spaces to align with fundamental computations. This improves generalization and reasoning in neural networks.
RAVEN\cite{zhang2019raven} combines both visual understanding and structural reasoning using a Dynamic Residual tree (DRT) graph developed from structural representation and aggregates latent features in a bottom-up manner. This provides a direction suggesting that augmenting networks with domain knowledge performs better than black-box neural networks. Scattering Compositional learner(SCL)\cite{SCL} presents an approach where the model learns a compositional representation by learning independent networks for encoding object, attribute representations and relationship networks for inferring rules, and using their composition to make a prediction. Our work bears similarity with this approach as both utilize background knowledge in composing a larger mechanism from elementary networks. CoPINet\cite{zhang2019learning} presents the Contrastive Perceptual Inference network which is built on the idea of contrastive learning, i.e. to teach concepts by comparing cases. The Dual-Contrast Network (DCNet)\cite{DCNet} works on similar lines as it uses 2 contrasting modules: rule contrast and choice contrast for its training. We draw inspiration from \cite{NAR_Analogies} which also presents a variation of the Neural Algorithmic Reasoning approach applied to visual reasoning.

\section{Conclusion}

In this work, we have proposed a novel neuro-symbolic reasoning approach where we learn neural-network based predicates operating on
a `symbolically-derived latent space' and use these in a symbolic search procedure to solve complex visual reasoning tasks, such as RAVENs Progressive Matrices.
Our experimental results (though preliminary, in that our predicates are composition of simple MLPs) indicate that our the approach points to a promising
direction for neuro-symbolic reasoning research.

\begin{acknowledgments}
This work is supported by “The DataLab” agreement between BITS Pilani, K.K.
Birla Goa Campus and TCS Research, India.
\end{acknowledgments}

\bibliography{main}

\appendix


\section{Overview of RAVEN dataset generation}

To give an overview of how the RAVEN dataset was generated, the authors used an A-SIG (Attributed Stochastic Grammar) to generate the structural representation of RPM. Each RPM is a parse tree that instantiates from this A-SIG. After this, rules and the initial attributes for that structure are sampled. The rules are applied to produce a valid row. This process is repeated 3 times to generate a valid problem matrix. The answer options are generated by breaking some set of rules. This structured representation is then used to generate images.

The RAVEN dataset provides a structural representation that is semantically linked with the image representation. The structural representation of the image space available in RAVEN makes it generalizable. As each image in a configuration follows a fixed structure, we use this knowledge to generate the corresponding symbolic representations. RAVEN has 10,000 problems for each configuration split into 6000: Train, 2000:Val and 2000:Test. We use the same split for training and validation and provide the results on the test set.

\section{Rule and Attribute definitions}



\subsection{Attributes}

\textbf{Number}: The number of entities in a given layout.
It could take integer values from [1; 9].\\
\textbf{Position}: Possible slots for each object in the
layout. Each Entity could occupy one slot.\\
\textbf{Type}: Entity types could be triangle, square,
pentagon, hexagon, and circle.\\
\textbf{Size}: 6 scaling factors uniformly distributed in
[0:4; 0:9].\\
\textbf{Color}: 10 grey-scale colors\\

\subsection{Rules}

4 different rules can be applied over rule-governing attributes. \\
\textbf{Constant}: Attributes governed by this rule would
not change in the row. If it is applied on Number or
Position, attribute values would not change across
the three panels. If it is applied on Entity level attributes, then we leave “as is” the attribute in each object across the three panels.\\
\textbf{Progression}: Attribute values monotonically increase
or decrease in a row. The increment or decrement
could be either 1 or 2, resulting in 4 instances in
this rule.\\
\textbf{Arithmetic}: There are 2 instantiations in this rule,
resulting in either a rule of summation or one of subtraction. Arithmetic derives the value of the attribute
in the third panel from the first 2 panels. For
Position, this rule is implemented as set arithmetics.\\
\textbf{Distribute Three}: This rule first samples 3 values
of an attribute in a problem instance and permutes
the values in different rows.

\section{Autoencoder, Neural Predicates and Image Encoder}

\subsection{Autoencoder}
\label{sec:autoencoder_loss}

The symbolic encoder $E^{\mathbf{S}}_{\theta}({s})$ is trained using the following losses as described in Section \ref{sec:approach}. As we want to reconstruct the multihot representation, a sum of negative-log likelihood is computed for each one-hot encoding present in the multihot representation. Here $p^k$ denotes the output nodes from the decoder corresponding to the $k^{th}$ attribute and $t^k$ denotes the one-hot input for the same attribute. In equations \ref{loss} and \ref{params} we use the example from RAVEN where the attributes are type, size, color, etc.
\begin{equation}
    \mathit{L}_\mathbf{S}(p,t) = \sum_{k \in \{type,size,col,\dots\}} - log(\frac{e^{p^{k}_{argmax(t^k)}}}{\Sigma_{i}e^{p^{k}_{i}}})
    \label{loss}
\end{equation}

\label{sec:img_encoder_los}
\begin{equation}
    \theta, \phi = argmin_{\theta,\phi}\Sigma_{s \in S}\mathit{L}_\mathbf{S}(\hat{s},s),  \text{where  }
    {\hat{s}}=D^{\mathbf{S}}_{\phi}(E^{\mathbf{S}}_{\theta}({s}))
    \label{params}
\end{equation}

\subsection{Neural Predicates for Rule Classification}
\label{C.2}
For every attribute, for each of its rule-type, we train an elementary neural network classifier to verify if the rule is satisfied in the row- this acts as our Rule Identification network. In this work, we refer to any rule identification network $F$ using a $(attribute, rule-type)$ pair for which it is trained. For rule-type Constant and Distribute Three we train a binary classifier. The rules of Progression and Arithmetic
are also associated with a value (e.g., Progression could have increments or decrements of 1 or
2), hence for rule-type Arithmetic and Progression, we train a multi-class classifier to also predict the value associated with the rule. Example: A neural network for Center: $F_{(Type, Constant)}$ is a binary classifier trained to identify if the row from the configuration Center has constant type (shape) across the 3 panels. Similarly a neural network for Left: $F_{(Size, Progression)}$ is a five-class classifier trained to classify if there is a progression in size in the Left component. This predicts 0 if there is no progression and predicts the progression value: increment or decrement (-2, -1, 1, 2) otherwise.

The latent representations $E^{\mathbf{S}}_{\theta}({s})$ obtained after encoding the symbolic representations of each of the three panels in the row are concatenated and sent as input to the neural networks. While training a neural net for a $(attribute,rule-type)$ pair, we categorize each row with the specific label for that particular rule and attribute, labeling it with 0 if the rule-type is not followed and with 1 or rule-value indicating the level of the rule-type when being followed.  As each of these rule-types is deterministic, we can obtain the rule-type and value for each row using their symbolic representations. The overview can be seen in component (ii) of Fig. \ref{pipeline} where we see the input for these elementary neural networks and the expected output determined for the $(attribute, rule-type)$ pair. With these labels, each network is optimized using cross-entropy loss. Each network is a shallow MLP classifier with 1 or 2 hidden layers whose dimensions are chosen depending on the configuration and validation set. These classifiers are trained using symbolic representations for each component across the various configurations and we provide the results in Table. \ref{F1-score_classification_networks}.

\subsection{Image Encoder}
To learn the latent representation of the unstructured data such that it mimics the symbolic latent space, we minimize the mean squared error over all pairs of symbolic and input representations obtained from $E^{\mathbf{S}}_{\theta}(s))$ and $E^{\mathbf{X}}_{\psi}(x)$ respectively. This enables us to use the previously learned neural networks for rule inference on our image data.

\begin{equation}
    \psi = argmin_{\psi}\Sigma_{(x,s)}{(E^{\mathbf{X}}_{\psi}(x)-E^{\mathbf{S}}_{\theta}(s))^2}
    \label{params_es}
\end{equation}

\begin{table}[h]
\caption{F1-score of rule classification networks. Note: Different components have different set of rules as in the case of Left-Right, Out-In Center, and Out-In Grid, wherein we train a separate set of networks for each component. Blank entries indicate that the rule setting does not exist for that particular component. Eg: Number attribute is always 1 in Center configuration.}
\label{F1-score_classification_networks}
\scriptsize
\begin{tabular}{@{}llccccccccccc@{}}
\toprule
 &  & Center & \multicolumn{2}{c}{Left-Right} & \multicolumn{2}{c}{Up-Down} & \multicolumn{2}{c}{Out-In Center} & 2x2 Grid & 3x3 Grid & \multicolumn{2}{l}{Out-In Grid} \\ \midrule
F1                   &              &  & Left & Right & Up & Down & Out & In &  &  & Out & In Grid \\ \midrule
\multirow{3}{*}{Typ} & Constant     & 1.0 & 1.0  &  1.0  & 1.0 & 1.0  & 0.99 &1.0 & 0.94 &0.94& 0.99 & 0.91    \\ \cmidrule(l){2-13}
                     & Distri Three & 1.0 & 0.99 & 0.99  &0.99& 1.0   & 0.99 &0.99&  0.92 &0.91&0.99 & 0.88    \\ \cmidrule(l){2-13}
                     & Progression  & 0.96 & 0.99 & 0.99  &0.99& 0.99 & 0.99 &0.99& 0.98 &0.96& 0.99 &   0.97    \\ \midrule
\multirow{4}{*}{Siz} & Constant     & 1.0 & 0.99 & 1.00  &0.99& 1.0  &  0.99 &1.0 &  0.91 &0.90& 1.0 &  0.95  \\ \cmidrule(l){2-13}
                     & Distri Three & 1.0 & 0.97 & 0.96  &0.98& 0.98 & 1.0  &0.98&  0.77 &0.72& 0.99 &  0.94   \\ \cmidrule(l){2-13}
                     & Progression  & 0.95  & 0.99 & 0.99  &0.99& 0.99 & 0.97 &0.99& 0.93  &0.94&0.96&  0.98   \\ \cmidrule(l){2-13}
                     & Arithmetic   & 0.91 & 0.96 & 0.96  &0.96& 0.97 & -  &0.96& 0.90 &0.84&   -  &    1.0  \\ \midrule
\multirow{4}{*}{Col} & Constant & 0.98  & 0.99 & 0.99  &0.99& 0.99 & -  &0.99& 0.82 &0.87&  -  &    0.86 \\ \cmidrule(l){2-13}
                     & Distri Three & 0.98 & 0.99 & 0.97  &0.98& 0.98 & - &0.98& 0.61 &0.74& - &    0.63 \\ \cmidrule(l){2-13}
                     & Progression  & 0.99 & 1.0 & 0.99  &0.99& 0.98 &   - &0.99& 0.95&0.92& - &    0.95 \\ \cmidrule(l){2-13}
                     & Arithmetic   & 0.93 & 0.91 & 0.93  &0.92& 0.95 &  -  &0.94& 0.78 &0.68&  -  &0.78 \\ \midrule
\multirow{4}{*}{Num} & Constant     & -  &  -   &   - &  - &  -   &  -  &  - &  0.93 &0.92& - &   0.96 \\ \cmidrule(l){2-13}
                     & Distri Three & -  &  -   &   -   &  - &  -   &  -  &  - &0.81&0.77&   - &    0.83 \\ \cmidrule(l){2-13}
                     & Progression  & -  &  -   &   -   &  - &  -   &  -  &  - &0.97&0.85&   - &    0.95 \\ \cmidrule(l){2-13}
                     & Arithmetic   & -  &  -   &   -   &  - &  -   &  -  &  - &0.96&0.84&   - &    0.94 \\ \midrule
\multirow{4}{*}{Pos} & Constant     & -  &  -   &   -   &  - &  -   &  -  & - & 0.93 &0.92&  -  &   0.96 \\ \cmidrule(l){2-13}
                     & Distri Three & -  &  -   &   -   &  - &  -   &  -  &  - &0.87&0.94&   - &    0.89 \\ \cmidrule(l){2-13}
                     & Progression  & -  &  -   &   -   &  - &  -   &  -  &  - &0.95&0.96&   - &    0.93 \\ \cmidrule(l){2-13}
                     & Arithmetic   & -  &  -   &   -   &  - &  -   &  -  &  - & 0.95 &0.92& -   &  0.93 \\

                     \bottomrule

\end{tabular}
\end{table}

\section{Search Algorithm}
\begin{algorithm}[H]
\caption{Search Overview}\label{alg:search}
\begin{algorithmic}
\footnotesize
\State $rules \gets []$
\State $e_{ij} \gets E_{\psi}^{\mathbf{X}}(x_{ij})$ \Comment{$x_{ij}$ refer to problem matrix images $(i:row,j:col)$}
\State $o_{k} \gets E_{\psi}^{\mathbf{X}}(y_{k})$ \Comment{$y_{k}$ refer to option images}
\State $R_1 \gets (e_{11},e_{12},e_{13}), R_2 \gets (e_{21},e_{22},e_{23})$
\For{$attr$ in attributes}
    \For{$rule$ in rule-types}
    \State $F \gets F_{(attr,rule)} $ \Comment{Use the trained neural network for the specific (attr,rule) pair}
    \State $p_1\gets F(R_1), p_2\gets F(R_2) $   \Comment{$p$ is the network output containing class wise probabilities}
    \If{$rule$ is Constant | Distribute Three}
        \If{$p_1 > \tau \wedge p_2 > \tau $}
            \State{$rules.add(attr,rule,1)$}
            \State{break}
        \EndIf
    \ElsIf{$rule$ is Progression | Arithmetic}
        \If{$argmax(p_1)!=0 \wedge argmax(p_2)!= 0 \wedge argmax(p_1)==argmax(p_2)$}
            \State $value \gets argmax(pred_1)$
            \State{$rules.add(attr,rule,value)$}
            \State{break}
        \EndIf
     \EndIf
\EndFor
\EndFor

\State $s_1,s_2,\dots,s_8 \gets 0$ \Comment{Initializing Scores for each option image}
\For{$k$ in \{1,2,\dots,8\}}
    \State{$o \gets o_k$}
    \For{($attr,rule,value$) in $rules$}
        \State{$R_3 \gets (e_{31},e_{32},o)$}
        \State $F \gets F_{(attr,rule)} $
        \State $p_3 \gets F(R_3) $
        \State $s_i = s_i +  p_{3,value}$ \Comment{Probability of the value inferred for the rule}
    \EndFor
\EndFor
\State{$ans \gets argmax(s_1,s_2,\dots,s_8)$}
\Return ans
\end{algorithmic}
\end{algorithm}

\end{document}